\documentclass[journal]{IEEEtran}

\ifCLASSINFOpdf
\else
   \usepackage[dvips]{graphicx}
\fi
\usepackage{url}

\usepackage[utf8]{inputenc} 
\usepackage[T1]{fontenc}
\usepackage{url}              
\usepackage{cite}             

\usepackage{graphicx}
 \usepackage{amsmath}
\usepackage{amssymb}
\usepackage{bm}
\usepackage{textcomp}
\usepackage{algorithmic}
\usepackage{algorithm}

\def\BibTeX{{\rm B\kern-.05em{\sc i\kern-.025em b}\kern-.08em
    T\kern-.1667em\lower.7ex\hbox{E}\kern-.125emX}}

\DeclareMathOperator*{\argmax}{arg\,max}

\usepackage{url}

\begin{document}

\title{Accelerating Convergence of Stein Variational Gradient Descent via Deep Unfolding}
\author{{Yuya Kawamura} and {Satoshi Takabe}
\thanks{Tokyo Institute of Technology 2–12–1 Ookayama, Meguro-ku, Tokyo 152–8550, Japan}
\thanks{This work was partly supported by JSPS Grant-in-Aid for Scientific Research: Grant Numbers 22H00514 and 22K17964.}}

\maketitle

\begin{abstract} 
Stein variational gradient descent (SVGD) is a prominent particle-based variational inference method used for {sampling a target} distribution. {SVGD has attracted interest for application in machine-learning techniques such as Bayesian inference.} In this paper, we propose {novel} trainable algorithms that incorporate a deep-learning technique {called} deep unfolding, into SVGD. This approach facilitates the learning of the internal parameters of SVGD, thereby accelerating its convergence speed. To evaluate the proposed trainable SVGD algorithms, we conducted numerical simulations of three tasks: sampling a one-dimensional Gaussian mixture, performing Bayesian logistic regression, and learning Bayesian neural networks. The results show that our proposed algorithms exhibit faster convergence than the conventional variants of SVGD.
\end{abstract}



\IEEEpeerreviewmaketitle

\section{Introduction}
\label{sec:introduction}
{B}{ayesian} {estimation is a probabilistic inference based on} the Bayesian interpretation of probability{\cite{bayes}}. Unlike the traditional maximum-likelihood estimation, Bayesian estimation maximizes the posterior probability calculated {using} the Bayes' theorem from {the likelihood and} prior distribution corresponding to {a hypothesis involving unknown variables}. 
This approach is a powerful means of expressing prior information in terms of probability and facilitates not only the estimation but also prediction of new data. 
However, the posterior distribution is generally difficult to evaluate analytically {because its normalization constant contains  multiple integrals of random variables.}
This computational difficulty has been a long-standing {issue} in  {various research fields such as machine learning and signal processing} {\cite{bayes2}}. 
{To circumvent it,} Markov-chain Monte-Carlo (MCMC) methods have been widely used in Bayesian estimation as approximate sampling methods for the posterior distribution\cite{mcmc}. {For instance, the Hamiltonian Monte Carlo (HMC) method enhances the sampling efficiency through physics-inspired proposals \cite{hmc}, while the no-U-turn sampler \cite{nuts} optimizes step selection in the HMC method. These  recent MCMC methods reduce the inherent computational complexity of Bayesian estimation.}
However, {MCMC methods still involve high computational costs because they serially update samples until the sample distribution is close to the target distribution.  
As a result, MCMC methods usually require long relaxation times, especially if the distribution lies in the high-dimensional space.}

Recently, particle-based variational inference (ParVI) has been proposed to improve the convergence speed and sampling efficiency. 
{Unlike MCMC methods,} ParVI approximates the distribution {based on the relative frequency distributions of}   
the particles representing the samples. 
{It updates} the particles to minimize the Kullback--Leibler (KL) divergence from the target distribution. Stein variational  gradient descent (SVGD)\cite{svgd} is a representative {ParVI algorithm, which uses a kernel function}. 
{The nonparametric nature of ParVI enhances its flexibility, resulting in particle efficiency better than that of the MCMC methods. In addition, ParVI algorithms, including SVGD, are closely related to  the Wasserstein  gradient flows \cite{parvis}.}
{A drawback of ParVI is its slow convergence speed, which depends on the initial and target distributions.
 In addition, minimizing the KL divergence is usually a nonconvex optimization.
 {The fact suggests} that gradient-based ParVI methods such as SVGD may converge to suboptimal solutions, leading to insufficient sampling of the target distribution. 
Although the performance of the algorithm depends on the choice of its internal parameters, the parameters are usually tuned in a heuristic manner.}

{The combination of deep learning and iterative algorithms has recently been attracting great interest. 
The deep-learning technique called deep unfolding (DU), in particular, enables the training of the internal parameters of an iterative algorithm, improving its convergence speed and performance}{\cite{du}}.
{In DU,} the iterative process of an algorithm is unfolded in the temporal direction and
{trainable parameters are embedded in the unfolded structure. 
These parameters are learned through standard deep-learning techniques such as back propagation and stochastic gradient descent if the unfolded algorithm consists of differentiable processes.}
This approach has been confirmed to enhance the convergence performance of the algorithms~\cite{Monga,Boyd} and has been applied in various fields associated with signal processing {\cite{tista,beam}}.

In this paper, {we examine the application of DU to ParVI to improve the convergence speed and sampling performance.
Specifically, we focus on SVGD and propose} a deep-unfolded SVGD (DUSVGD) method, whose internal {step-size} parameters are learnable. 
Furthermore, inspired by the Chebyshev step\cite{che2}, which is a step-size sequence for gradient descent, we propose a {Chebyshev step-based DUSVGD (C-DUSVGD) algorithm. 
The number of trainable parameters of C-DUSVGD is only two, whereas that of DUSVGD is a constant {with respect to} the number of particles and dimension of variables. }
Finally, we numerically {examine} the performance of the proposed methods. Namely, we perform three different tasks: sampling a multimodal distribution, performing Bayesian logistic regression, and learning a Bayesian neural network.

The {remainder} of this paper is organized as follows. Section \MakeUppercase{\romannumeral 2} briefly reviews SVGD, DU, and the Chebyshev step. Section \MakeUppercase{\romannumeral 3} describes DUSVGD and C-DUSVGD and Sec. \MakeUppercase{\romannumeral 4} evaluates their performance. Section \MakeUppercase{\romannumeral 5} concludes the paper.

\section{Preliminaries}
{In this section, we briefly review the previous studies related to this paper.}

\subsection{Stein Variational Gradient Descent}\label{sec_SVGD}
SVGD is a {ParVI} algorithm that estimates a {target} distribution by minimizing the KL divergence between the distribution of particles in SVGD and the {target} distribution. Here, we briefly describe the SVGD algorithm\cite{svgd}.

SVGD uses a set of particles $\{\bm x_i\}_{i=1}^M$ to approximate the {$d$-dimensional target} distribution $p(\bm x)$. The empirical distribution of the particle set is represented by $q(\bm x) = M^{-1}\sum_{i=1}^M \delta(\bm x - \bm x_i)$ ($\bm x_i\in\mathbb R^d, i\in [M]:=\{1,\dots,M\}$), where $\delta(\cdot)$ is the Dirac delta function. 
The particles are independently and identically distributed {random variables} from an initial distribution $q_0(\bm x)$, and are iteratively updated according to the following equation:
\begin{align}
\bm x^{(t+1)} =\bm x^{(t)} + \epsilon^{(t)} \bm\phi(\bm x^{(t)}), \label{tra}
\end{align}
where $\bm\phi$ represents the vector-valued function indicating the direction of particle update and $\epsilon^{(t)} (>0)$ is the step size of the $t$-th iteration ($t=0,1,\dots$). 
The distribution followed by (\ref{tra}) is denoted as $q_{[\epsilon \bm \phi]}$. 
{In SVGD, $\bm\phi$ {is set to} the direction that minimizes the KL divergence $\mathrm{KL} (q_{[\epsilon \bm \phi]} || p)$. By setting $\mathcal{F}$ as the function space to which $\bm \phi$ belongs, the optimal update direction is given by 
\begin{align}
\bm \phi^* = \argmax_{\bm \phi \in \mathcal{F}} \left\{\left.-\frac{d}{d\epsilon} \mathrm{KL}(q_{[\epsilon \bm \phi]} || p)\right|_{\epsilon=0} \right\} = \mathbb{E}_{\bm x \sim q}[\mathcal{A}_p\bm \phi(\bm x)], 
\end{align}
{where} $\mathcal{A}_p$ is the Stein operator, defined as $\mathcal{A}_p \bm\phi(\bm x)=\nabla_{\bm x}\log p(\bm x) \bm\phi(\bm x) ^\top + \nabla_{\bm x} \bm\phi(\bm x)$. This optimization problem can be solved by restricting $\mathcal{F}$ to a unit ball of a vector-valued reproducing kernel Hilbert space $\mathcal{H}^d$ defined by a positive definite kernel $k(\cdot,\cdot)$. {Then, the optimal direction} is given by  
\begin{align}
\bm \phi^*(\cdot) \propto \mathbb{E}_{\bm x \sim q}[\mathcal{A}_p k(\bm x,\cdot)]. \label{eq_svgd_grad}
\end{align}
Consequently, {SVGD is summarized to Alg.~\ref{algsvgd} as an iterative algorithm}. 

\begin{algorithm}[t]
\caption{Stein variational gradient descent \cite{svgd}}
\label{algsvgd}
\begin{algorithmic}[1]
\REQUIRE Initial set of particles $\{\bm x_i^{(0)}\}_{i=1}^M$
\ENSURE Set of particles $\{\bm {\hat x}_i\}_{i=1}^M$ approximating the target distribution $p(\bm x)$
\FOR{$t= 0,1 \dots T-1$}
\STATE $\displaystyle \bm x_i^{(t+1)} \leftarrow \bm x_i^{(t)}+\epsilon^{(t)}\bm\phi(\bm x_i^{(t)})$ \\
$\displaystyle \bm\phi(\cdot)\!=\!\frac{1}{M} \sum_{j=1}^M \left[k(\bm x_j^{(t)},\cdot)\nabla_{\bm x_j^{(t)}}\log p(\bm x_j^{(t)})\!+\!\nabla_{\bm x_j^{(t)}}k(\bm x_j^{(t)},\cdot)\right]$
\ENDFOR
\STATE $\{ \bm {\hat x}_i \}_{i=1}^M \leftarrow \{ \bm x_i^{(T)}\}_{i=1}^M$
\end{algorithmic}
\end{algorithm}

In \cite{svgd}, the radial basis function (RBF) kernel is used as the positive definite kernel. {The RBF kernel is defined by}
\begin{align}
k(\bm x,\bm x') = \exp \left(-\frac{1}{h} \|\bm x-\bm x'\|_2^2 \right), \ h = \frac{\mathrm{median}(\{ \bm x_i\}_{i=1}^M)}{\log M}.\label{karnel}
\end{align}
The parameter $h$ aims to stabilize the computation of the gradient  (\ref{eq_svgd_grad}) by ensuring that $\sum_j k(\bm x_i , x_j) \approx M \exp(-\mathrm{median}^2/h) = 1$. 
{In this paper, we use the RBF kernel (\ref{karnel}) unless otherwise noted}.

{The drawback of SVGD is that the minimization of the KL divergence is a nonconvex optimization, in general, leading to convergence to a suboptimal distribution of particles. 
Further, {the difficulty of sampling multinomial distributions with distant peaks depends} on an initial distribution, using SVGD }{\cite{mpsvgd,asvgd}}. 
{In addition, the sampling and convergence performance apparently depend on the step-size parameter $\{\epsilon^{(t)}\}$, which is usually chosen heuristically.}

\subsection{Deep Unfolding}\label{sec_DU}
As described in Sec. \ref{sec:introduction}, DU is a deep-learning technique that unfolds {an existing iterative} algorithm and {trains} its internal parameters {to improve its performance}\cite{du}. 

We explain the DU of gradient descent as an example. Let the objective function be $f : \mathbb{R}^n \rightarrow \mathbb{R}$. The {update rule of} gradient descent that minimizes the function, with the maximum number of iterations $T$, is given by
\begin{align}
\bm x^{(t+1)} = \bm x^{(t)} - \epsilon^{(t)} \nabla f(\bm x^{(t)}) \quad (t = 0, 1,\dots , T-1),
\end{align}
{where $\bm \epsilon:=(\epsilon^{(0)},\dots,\epsilon^{(T-1)})$ represents the step-size parameters. 
In DU, we consider learning $\bm \epsilon$ through supervised learning.}
{Let} a minibatch $\mathcal{B}$ be given by
\begin{align}
\mathcal{B} := \left\{ (\bm x_1^{(0)}, \bm x_1^\ast), (\bm x_2^{(0)}, \bm x_2^\ast), \dots , (\bm x_B^{(0)}, \bm x_B^\ast) \right\},
\end{align}
{where} $B$ is the minibatch size, {$\bm x_b^{(0)}$ ($b\in [B]$) is the initial point of gradient descent, and $\bm x_b^\ast$ is the corresponding optimal solution}. 
{During $E$ learning epochs, the trainable step sizes are optimized by minimizing a proper loss function $L^{(t)}_\mathcal{B}(\bm \epsilon)$ {such as the mean-squared error}  {between the output of gradient descent $\bm{x}^{(t)}$ and optimal solution $\bm x^\ast$. The training process is summarized in Alg.~\ref{algdugd}. } 

\begin{algorithm}[t]
\caption{{Incremental training of step sizes in deep-unfolded gradient descent}}
\label{algdugd}
\begin{algorithmic}[1]
\REQUIRE Initial set of step sizes $\{\epsilon^{(t)} \}_{t =0}^{T-1}$
\FOR{$t' = 0,1 ,\dots ,T-1$}
\FOR{$\mathrm{epoch} = 1, \dots ,E$}
\STATE Generate a random minibatch $\mathcal{B}$
\STATE Compute the loss function $L^{(t')}_\mathcal{B}(\bm \epsilon)$
\STATE Compute the gradient vector of the loss function: $\bm G := \nabla L^{(t')}_\mathcal{B}(\bm \epsilon)$
\STATE Update $\{ \epsilon^{(i)} \}_{i=0}^{t-1}$ using $\bm G$ and an optimizer
\ENDFOR
\ENDFOR
\end{algorithmic}
\end{algorithm}

{In Alg. \ref{algdugd}, we employ the so-called incremental training to avoid gradient vanishing}{\cite{tista}}. {In the training, we first train the unfolded algorithm for $t'=1$ iteration. After $E$ epochs are fed, the number of iterations $t'$ is set to two.  Similarly, all parameters are trained in an incremental manner. Note that back propagation is applicable to learning parameters because all processes of the algorithm are differentiable, leading to a fast training process.}

\subsection{Chebyshev Step}\label{sec_Cheb}

The Chebyshev step is a step-size {sequence that accelerates the convergence speed of gradient descent  \cite{che}. The sequence is theoretically derived to understand the trained step-size parameters of the deep-unfolded gradient descent. Here, we briefly review the Chebyshev step.} 
Let $\bm A \in \mathbb{R}^{n \times n}$ be a positive definite symmetric matrix and {consider minimizing} $f(\bm z)=\bm z^\top \bm A \bm z/2$. 
{The update rule of} gradient descent with step sizes $\{ \tilde \epsilon^{(t)}\}_{t=0}^{T-1}$ is given by
$\bm z^{(t+1)} = (\bm I_n - \tilde \epsilon^{(t)}\bm A) \bm z^{(t)} : = \bm W^{(t)} \bm z^{(t)}$. 
{Let us assume that $T$ step sizes are trained and that the algorithm is executed by setting $\epsilon^{(t)} = \tilde \epsilon^{(u)}$ with $t \equiv u\ (\mathrm{mod}\, T)$, where $\{\tilde \epsilon^{(u)}\}_{u=0}^{T-1}$ represents the learned step sizes.
Then, the update rule for $\bm z$ in every $T$ iterations is given by  
$\bm z^{((k+1)T)} = \left(\prod _{t=0}^{T-1} \bm W^{(t)}  \right)  \bm z^{(kT)} := \bm Q^{(T)} \bm z^{(kT)}$ ($k=0,1,\dots$).
Recall that the matrix $\bm Q^{(T)}$ is a function of  $\{\tilde \epsilon^{(u)}\}_{u=0}^{T-1}$. 
The Chebyshev step is obtained as a solution that minimizes the upper bound of the spectral radius 
$\rho (\bm Q^{(T)})$, which relates to the convergence speed of gradient descent. 
Finally, the Chebyshev step is obtained as a reciprocal of the root of the Chebyshev polynomial of order $T$. 
Assuming that the maximum and minimum eigenvalues of $\bm A$ are $\lambda_n$ and $\lambda_1$, respectively, 
the Chebyshev steps of length $T$ \cite{che} are given by  
\begin{equation}
\epsilon_c^{(t)} = \left[ \frac{\lambda_n + \lambda_1}{2}+ \frac{\lambda_n - \lambda_1}{2} \cos \left( \frac{2t+1}{2T}\pi \right) \right]^{-1},\label{eq_chev}
\end{equation} 
for $t = 0,1,\dots,T-1$.}  
{The convergence rate of gradient descent with the Chebyshev step is proved to be smaller than that of the gradient descent with a  constant step size, indicating that the Chebyshev step can accelerate the convergence speed as well as a deep-unfolded gradient descent.}

\section{Proposed Method}

{As described in the last part of Sec. \ref{sec_SVGD}, SVGD is regarded as a gradient descent that minimizes the KL divergence, which is a nonconvex function in general. 
Here, we propose novel DU-based SVGD algorithms that allow us to learn the step sizes {adapted to} a target distribution or dataset. }

\subsection{DUSVGD}

\begin{figure}[t]
\centering
\includegraphics[width=0.95\columnwidth]{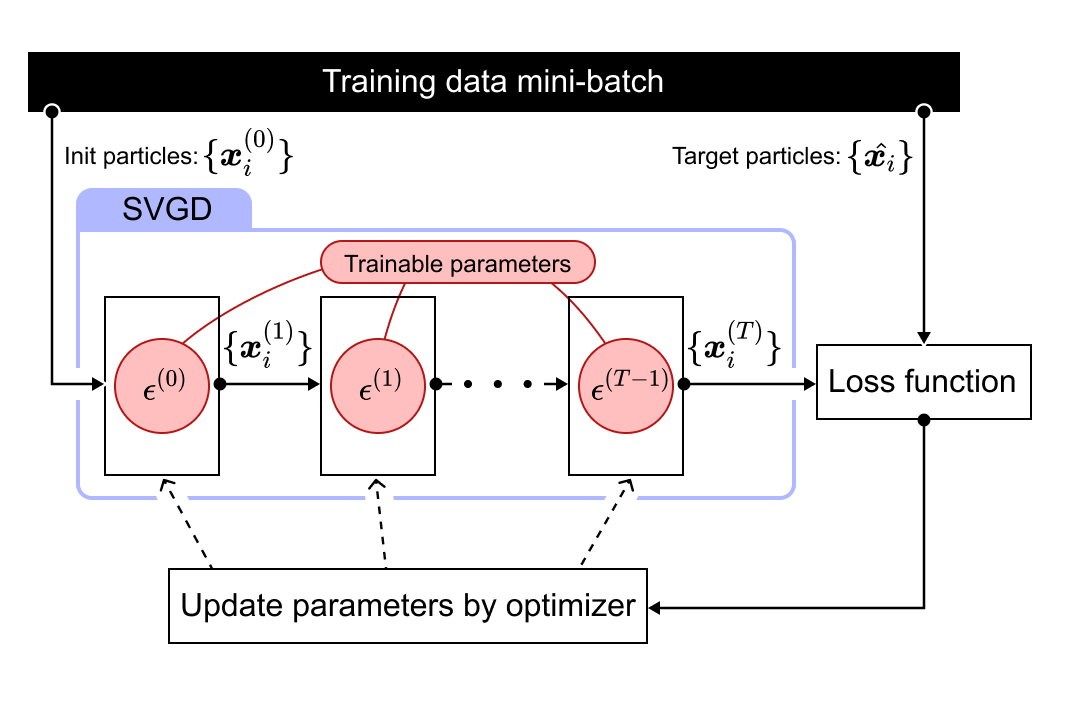}
\caption{Structure of DUSVGD.}
\label{duf}
\end{figure}

The convergence performance of SVGD depends on the step sizes $\{ \epsilon^{(t)}\}$. Similar to the deep-unfolded  gradient descent, we consider making the step sizes learnable by applying DU to SVGD. 
{Namely, in the update rule of} particles $\{ \bm  x_i\}_{i = 1}^M$, 
\begin{equation}
	 \bm x_i^{(t+1)} = \bm  x_i^{(t)}+\epsilon^{(t)} \bm\phi(\bm x_i^{(t)}), \label{eq_dusvgd}
\end{equation}
{where $\{\epsilon^{(t)}\}_{t=0}^{T-1}$ are trainable step-size parameters for a given $T$.
In this paper, the trainable SVGD  is called DUSVGD.
Note that all particles share the same step size for each $t$, although we can train $\epsilon^{(t)}_i$ according to the particle. 
This is because we make the number of trainable parameters  constant with respect to the number of particles, $M$, leading to a  scalable and fast training process. 
The value of $T$ is fixed in advance. In the following, we repeatedly use the trained step sizes for $t>T$.
Namely, we set $\epsilon^{(t')}=\epsilon^{(t)}$ with $t'\equiv t$ (mod $T$) for any $t'=0,1,\dots$.
}

{To prevent gradient vanishing in the training process of SVGD, the incremental training described in Sec.\ref{sec_DU} is used.
The architecture and training process of DUSVGD are depicted in {Fig.} \ref{duf}. 
Note that we use a proper loss function depending on the task of DUSVGD, which is denoted in Sec.~\ref{sec_res}.}

Note that the proposed DUSVGD is different from nueral variational gradient descent (NVGD)~\cite{NVGD} that learns the function $\phi$ as a neural network. Although NVGD is more flexible than DUSVGD, the number of trainable parameters of NVGD is larger than that of DUSVGD, implying that the training cost of NVGD is more costly.

\subsection{C-DUSVGD}

{Let us consider a simple case where an SVGD with $M=1$ particle attempts to sample a target $d$-dimensional Gaussian distribution with zero mean and variance-covariance matrix $\bm \Sigma$. 
Then, the update rule (\ref{tra}) with the RBF kernel is rewritten as
$\bm x_1^{(t+1)} = \bm  x_1^{(t)}-\epsilon^{(t)}\bm \Sigma^{-1}\bm x_1^{(t)}$
because $k(\bm x,\bm x)=1$ and $\nabla_{\bm x}k(\bm x,\bm y)=0$ hold when $\bm y=\bm x$.}
{This indicates that SVGD is equivalent to a gradient descent that minimizes a quadratic function, in this case. 
Notice that the setting is the same as that for the Chebyshev step, suggesting that the Chebyshev step gives  (sub)optimal step sizes.} 

{From this observation, we consider applying the Chebyshev step to SVGD. 
However, the Chebyshev step requires $\lambda_1$ and $\lambda_n$, the minimum and maximum eigenvalues of the Hessian matrix at the solution for general KL divergence, which is computationally difficult to estimate. 
We alternatively consider training $\lambda_1$ and $\lambda_n$ based on DU. Namely, the update rule for particles $\{ \bm  x_i\}_{i = 1}^M$ is written as}
\begin{align}
	\lambda_1 &= \alpha^2, \label{la1}\\
	\lambda_n &= \alpha^2 + \beta^2,  \label{lan}\\
	\epsilon_c^{(t)} &= \left[ \frac{\lambda_n + \lambda_1}{2}+ \frac{\lambda_n - \lambda_1}{2} \cos \left( \frac{2(T-t)-1}{2T}\pi \right) \right]^{-1},  \label{che2}\\
	  \bm x_i^{(t+1)} &= \bm  x_i^{(t)}+\epsilon_c^{(t)} \bm\phi(\bm x_i^{(t)}),
 	\end{align}
{where} $\alpha$ and $\beta$ are trainable parameters. Equation (\ref{che2}) represents the Chebyshev step\footnote{The order of the Chebyshev step affects the performance of C-DUSVGD in general. In this paper, we use the Chebyshev step in the reverse order, which exhibits better performance than (\ref{eq_chev}).}, and parameters are defined by (\ref{la1}) and (\ref{lan}) to satisfy the condition $0<\lambda_1 \leq \lambda_n$. 
This trainable SVGD is named C-DUSVGD, in this paper. 
C-DUSVGD {is advantageous when compared to DUSVGD in terms of the lower training cost because} {it has only two} {trainable parameters for any $M$ and $T$, whereas DUSVGD is more flexible than C-DUSVGD for approximating target distributions.}

{In the training process of C-DUSVGD, we do not require incremental learning because gradient vanishing is not observed. C-DUSVGD is trained only by the output after $T$ iterations.
The trained step sizes are used periodically for $t>T$.}
 
\section{Numerical Experiments}\label{sec_res}

To evaluate the performance of the proposed {trainable SVGD algorithms}, we {executed numerical simulations} {of} three different tasks: sampling a one-dimensional Gaussian mixture distribution, Bayesian logistic regression, and learning a Bayesian neural network. 
{Trainable SVGD algorithms were implemented using} {PyTorch 2.0.0 \cite{pytorch}} {and learned using the Adam optimizer}{\cite{adam}.}
For comparison, we used SVGD with RMSProp {\cite{rmsprop}}, a commonly used optimization method, and SVGD with a fixed step size. 
Note that other samplers such as the HMC were omitted because their performance was overwhelmed by that of the SVGD~\cite{svgd}.

\subsection{{Sampling of} Gaussian Mixture Distribution}

\begin{figure}[t]
  \centering
  \includegraphics[width=0.95\columnwidth]{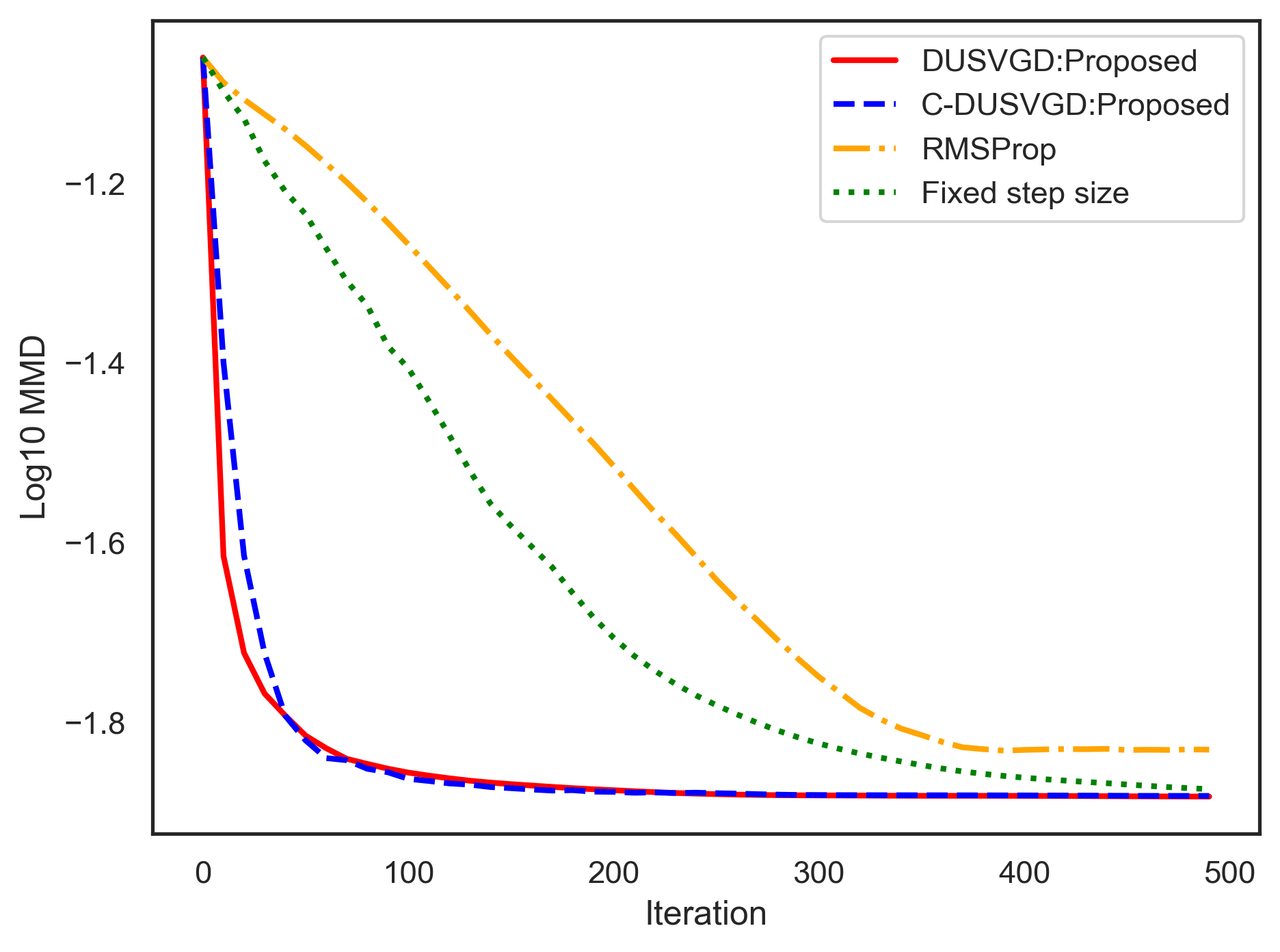}
  \caption{Dependency of MMD on the number of iterations for DUSVGD, C-DUSVGD, SVGD with RMSProp, and SVGD with a fixed step size in sampling a one-dimensional Gaussian mixture distribution.}
  \label{fig:gaussian}
\end{figure}

As a simple example, we applied SVGD to approximate a one-dimensional Gaussian mixture distribution. The probability density function of the target distribution $p(x)$ was given by 
\begin{align}
p(x) = \frac{3}{4} \mathcal{N} (x; -2, 1) + \frac{3}{4}\mathcal{N}(x; 2.5, 1), \label{px}
\end{align}
where $\mathcal{N}(x;\mu,\sigma^2)$ denoted the probability density function of the Gaussian distribution with mean $\mu$ and variance $\sigma^2$.
The initial distribution of particles in the SVGD was set to $q_0(x) = \mathcal{N}(x; -2,1)$, with the number of particles $M=100$. 
The maximum number of iterations for {training (C-)DUSVGD} was set to $T=10$, and the trained step sizes were used periodically. 
The {dataset consisted of} 1000 random numbers generated from (\ref{px}), with 90\% used for training and 10\% for testing. 
The minibatch size was $B=50$, with $E=10$ learning epochs for DUSVGD and $E=40$ for C-DUSVGD. 
{As a loss function, we used the maximum mean discrepancy (MMD) {\cite{mmd}}. MMD estimated the distance between two distributions $p$ and $q$ using sample sets $X = \{x_1,\dots,x_m\}$ and $Y=\{y_1,\dots,y_n\}$ independently drawn from $p$ and $q$, respectively. It was defined by}
\begin{equation}
\mathrm{MMD}[\mathcal{F}',X, Y] 
= \langle X,X \rangle -2  \langle X,Y \rangle  + \langle Y,Y \rangle,  
\end{equation}
where $\mathcal{F}'$ represented the unit ball in the reproducing kernel Hilbert space corresponding to the positive definite kernel $k'(x,x')$, and 
\begin{equation}
\langle Z,W \rangle
= \frac{1}{m'n'}\sum_{i=1}^{m'}\sum_{j=1}^{n'} k'(z_i,w_j),
\end{equation}
where $Z=\{z_1,\dots,z_{m'}\}$ and $W=\{w_1,\dots,w_{n'}\}$ were sample sets.
 In this experiment, the RBF kernel $k'(x,x') = \exp (-||x-x'||^2 /2) $ was used for MMD
 The learning rate for the Adam optimizer was set to $1.0\times 10^{-2}$ for DUSVGD and $1.0\times 10^{-3}$ for C-DUSVGD. 
 The initial value of trainable parameters were $\gamma_t = 2.0$ for DUSVGD and $(\alpha,\beta)=(0.3,1.0)$ for C-DUSVGD.

Figure \ref{fig:gaussian} shows the {MMD as a function of} the number of iterations for DUSVGD, C-DUSVGD, SVGD with RMSProp (referred to as ``RMSProp''), and SVGD with a fixed step size (referred to as ``fixed step size'').
MMD was calculated using randomly chosen samples and averaged over $50$ trials. 
In terms of the convergence performance, DUSVGD and C-DUSVGD, which learned the step sizes, converged faster than RMSProp and fixed step size. 
For example, the number of iterations required to achieve $\log_{10}\mathrm{MMD}=1.7\times 10^{-1}$ was approximately 20 for DUSVGD, 30 for C-DUSVGD, 280 for RMSProp, and 200 for fixed step size. 
This shows the effectiveness of learning the step size parameters of SVGD. 
The superiority of  DUSVGD over C-DUSVGD suggests that the flexibility of the architecture affects the convergence performance.

Figure \ref{fig:gaussian_d} shows the distribution of particles of SVGD {algorithms after $100$ iterations. 
We applied kernel density estimation using the RBF kernel of bandwidth $0.2$} to estimate the particle distribution. 
{We found that} DUSVGD and C-DUSVGD could approximate the distribution more accurately with fewer iterations than SVGD with  RMSProp and fixed step size

.

\begin{figure}[t]
  \centering
  \includegraphics[width=0.95\columnwidth]{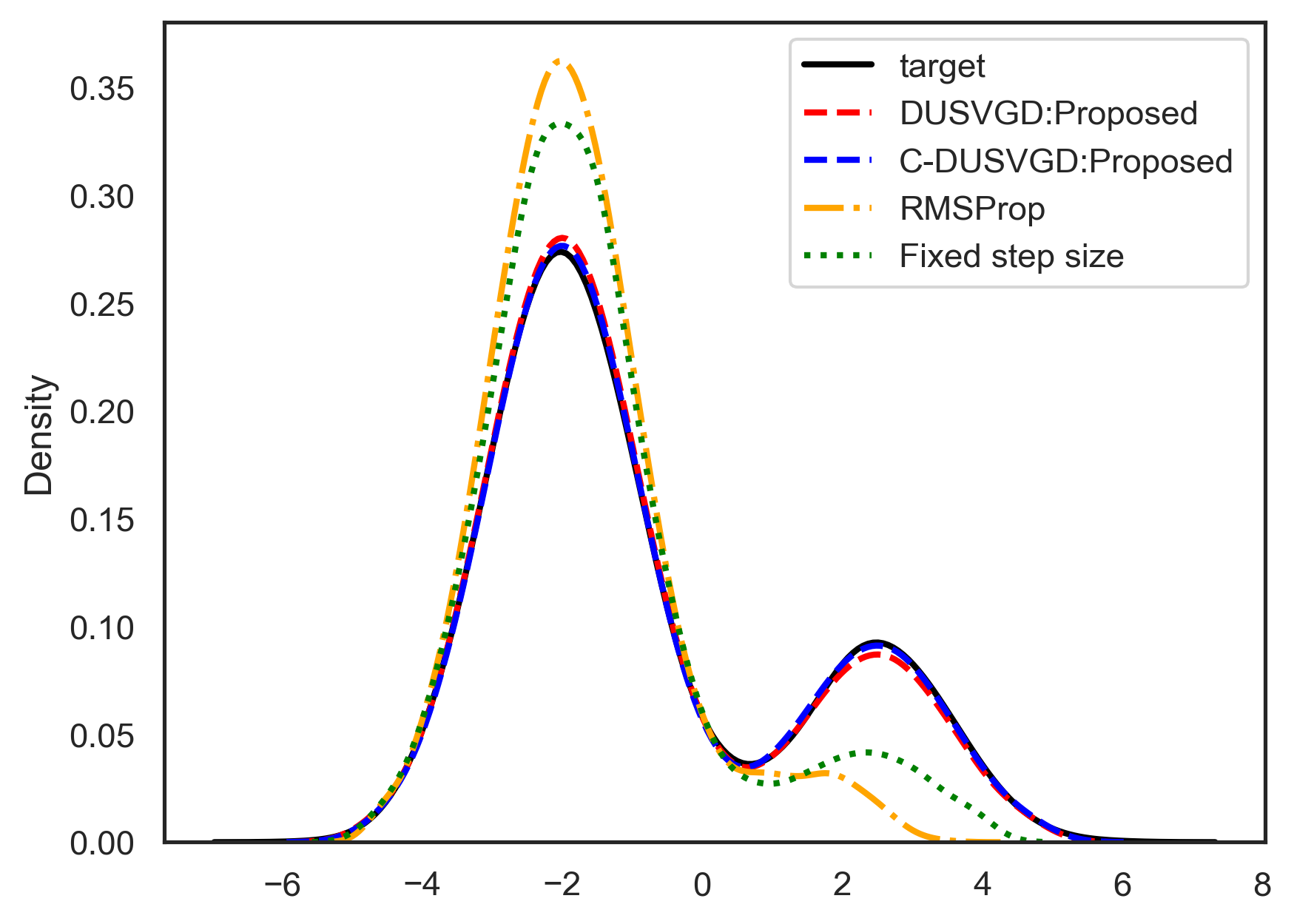}
  \caption{Distributions of particles of (C)-DUSVGD and SVGD algorithms after $100$ iterations in approximating a one-dimensional Gaussian mixture distribution.
  Each distribution is obtained via kernel density estimation with an RBF kernel of bandwidth $0.2$. 
  The target distribution (\ref{px}) is represented by the solid line.}
  \label{fig:gaussian_d}
\end{figure}

\subsection{Bayesian Logistic Regression}

Next, we consider a more practical problem using real data. Here, we apply SVGD to the Bayesian logistic regression for a binary classification dataset\footnote{\url{https://www.csie.ntu.edu.tw/~cjlin/libsvmtools/datasets/binary.html}}. The observational data $D = \{(\bm x_n,t_n)\}_{n=1}^N$ consists of class label $t_n \in \{-1,1\}$ and $K$ explanatory variables $\bm x_n \in \mathbb{R}^K$ for each data point, where $N=581012$ and $K=54$. 
In the experiment, the target distribution for the regression coefficients $\bm w \in  \mathbb{R}^K$ is {modeled as a posterior distribution used in the previous study \cite{blr}. 
The target distribution is given by}
\begin{align}
p(\alpha) &= \mathrm{Gamma}(\alpha;a,b), \label{alpha}\\
p(w_k|\alpha) &= \mathcal{N}(w_k;0,\alpha^{-1}),\quad (k \in [K]) \label{kaiki}\\
p(t_n = 1|\bm x_n, \bm w)&= \frac{1}{1+\exp(-\bm w^\top \bm x_n)},\quad (n \in [N]) \label{logi}
\end{align}
where {$\mathrm{Gamma}(\alpha;a,b)$ is a Gamma distribution defined by}
\begin{align}
\mathrm{Gamma}(\alpha;a,b)  &= \left\{
\begin{array}{ll}
\displaystyle \frac{1}{b^a \Gamma(a)} \alpha ^ {a-1} e^{-\frac{\alpha}{b}}& (\alpha \geq 0)\\
0 & (\alpha \leq 0)
\end{array}
\right. ,\\
 \Gamma(a) &= \int _0^\infty x^{a-1} e^{-x} dx .
\end{align}
{In the model, $\alpha$ represents a precision parameter following the Gamma distribution with hyperparameters  $a$ and $b$.} As a Bayesian estimation, (\ref{alpha}) and (\ref{kaiki}) represent the prior distributions and {$p(\bm{t}|\bm{X},\bm w,\alpha)$} corresponds to the likelihood. {Here, $\bm{t} = (t_1, ..., t_N)^\top$ represents a vector of all class labels, while $\bm{X} = (\bm{x}_1, ..., \bm{x}_N)$ denotes a matrix comprising all observed data points.}
SVGD approximates the target posterior distribution $p(\bm w, \alpha |D)$. 
The values of hyperparameters are set to $a =1$ and $b=0.01$.

{In SVGD, each particle was a $(K+1)$-dimensional random variable containing the values of $\bm w$ and $\alpha$. 
In the experiment, $M=100$ particles were used, which were initialized by prior distributions.}  
The number of iterations for {training (C-)DUSVGD} was set to $T=10$, and the learned step sizes were used periodically. 
{80\% of data points were used for training and the remaining}  {20\%} {for testing.} The minibatch size was $B=1$, with $E=500$ learning epochs for DUSVGD and $E=1000$ for C-DUSVGD. 
{The cross-entropy function between the true and estimated class labels was used as the loss function 
because the true posterior distribution was unknown in this case.
Thus, the loss function is defined by}
\begin{align}
H (\bm t, \bm{\hat{t}})= -\frac{1}{N_{train}}\sum_{n=1}^{N_{train}} \left(t_i \log \hat{t}_i + (1-t_i)\log(1-\hat{t}_i) \right),
\end{align}
where $N_{train}$ is the number of training data, $t_i$ is a true class label in training data, and $\hat{t}_i$ is the corresponding estimated class label.
 The learning rate for the Adam optimizer was set to $1.0\times 10^{-7}$ for DUSVGD and $1.0\times 10^{-1}$ for C-DUSVGD. 
 The initial value of trainable parameters were $\gamma_t = 2.0\times 10^{-5}$ for DUSVGD and $(\alpha,\beta)=(200,500)$ for C-DUSVGD.

\begin{figure}[t]
  \centering
  \includegraphics[width=0.95\columnwidth]{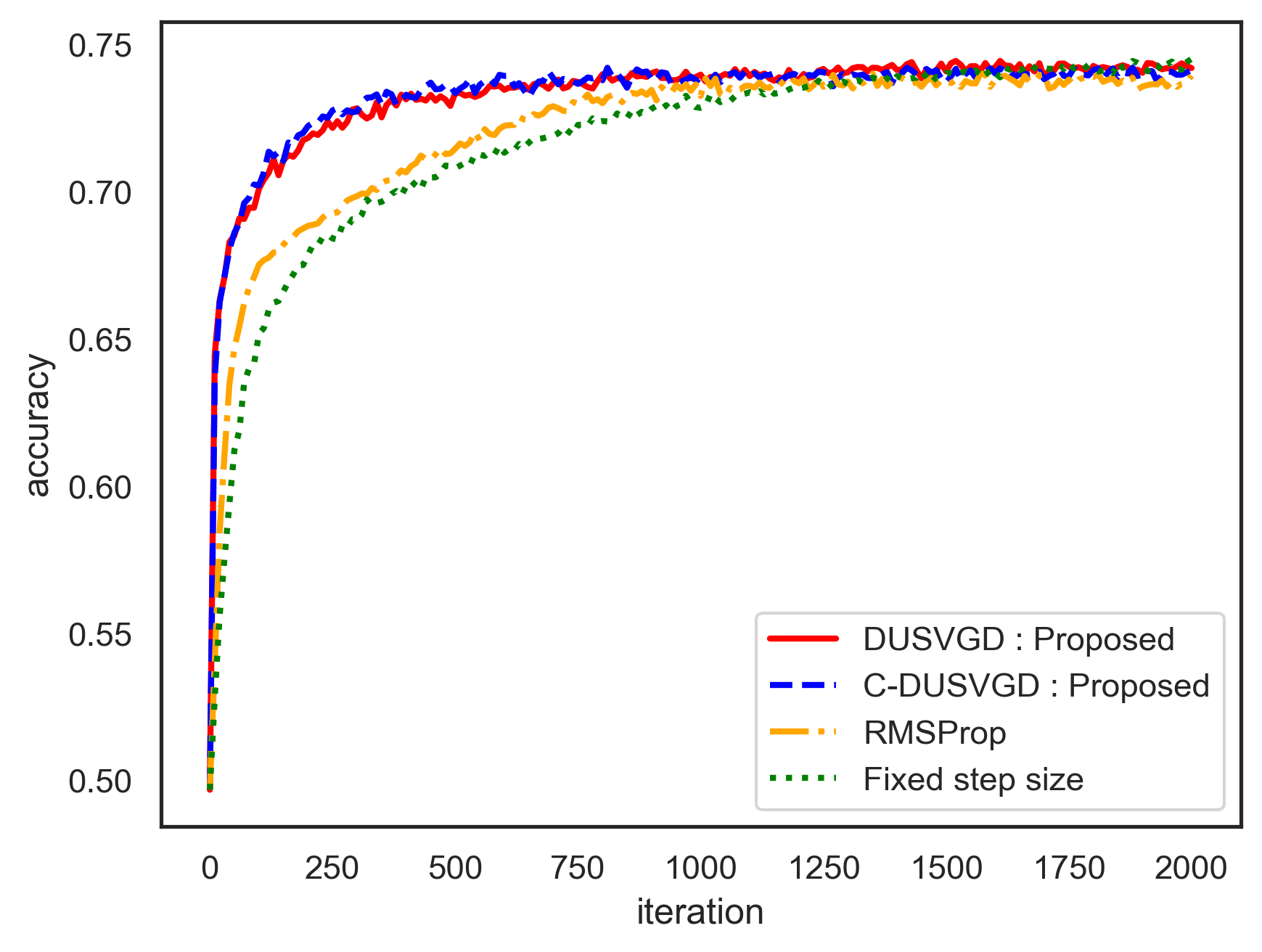}
  \caption{Dependency of accuracy on the number of iterations for DUSVGD, C-DUSVGD, RMSProp, and fixed step size in the Bayesian logistic regression problem.}
  \label{fig:blr}
\end{figure}

Figure \ref{fig:blr} shows the dependency of the accuracy on the number of iterations for DUSVGD, C-DUSVGD, RMSProp, and fixed step size in the Bayesian logistic regression problem. 
{In the figure, the accuracy of the label prediction was averaged  over 30 trials.} 
In terms of convergence performance, it was observed that DUSVGD and C-DUSVGD showed reasonable improvement compared to the other SVGD. For instance, the accuracy at 500 iterations was approximately 73.3\% for DUSVGD, 73.4\% for C-DUSVGD, 71.6\% for RMSProp, and 70.7\% for fixed step size.
The result shows that the proposed (C-)DUSVGD exhibits  reasonable performance gain compared to the existing SVGD, even for Bayesian logistic regression with real data.  
It is emphasized that the training was executed with only $T=10$ iterations, resulting in an efficient training cost.

\subsection{Learning Bayesian Neural Networks}

As the second numerical experiment with real data, we  {examined} the application of learning Bayesian neural networks for a regression dataset\footnote{\url{https://www.cs.toronto.edu/~delve/data/boston/bostonDetail.html}}. 
The training data $D=(\bm X,\bm y)=(\{\bm x_n\}_{n=1}^N,\{y_n\}_{n=1}^N)$ consisted of a target value $y_n\in \mathbb{R}$ and corresponding $K$-dimensional input vectors $\bm x_n\in \mathbb{R}^K$, {where $N=506$ and $K=14$}.
In the experiments, we assumed that the target values were obtained as $y_n=f(\bm x_n;\mathcal{W})+\eta_n$, where $f(\cdot;\mathcal{W})$ was a two-layer fully connected neural network with 50 units, ReLU activation function in the middle layer, {and identity activation function in the output layer}. 
{Unlike typical} {neural networks, the model contained a Gaussian noise $\eta_n$ following 
$\mathcal{N}(0,\gamma^{-1})$. 
Our task was to learn the weight matrix $\mathcal{W}=(\bm w_1,\dots,\bm w_{50})^\top$ as a Bayesian neural network. Namely, we assumed the weights were random variables and introduced the following probabilistic model.}
\begin{align}
  p(\lambda) &= \mathrm{Gamma}(\lambda;a_{\lambda},b_{\lambda}),\label{lambda}\\
  p(\gamma) &= \mathrm{Gamma}(\gamma;a_{\gamma},b_{\gamma}),\label{gamma}\\
  p(\mathcal{W}|\lambda) &= \prod_{j=1}^{50}\mathcal{N}(\bm w_{j}|0,\lambda^{-1}),\label{weight}\\
  p(\bm y|\bm X,\mathcal{W},\gamma) &= \prod_{n=1}^{N}\mathcal{N}(y_n|f(\bm x_n;\mathcal{W}),\gamma^{-1})\label{y},
\end{align}
where $\lambda$ and $\gamma$ were precision parameters following  Gamma distributions, and $a_{\lambda},b_{\lambda},a_{\gamma}$, and $b_{\gamma}$ were the hyperparameters of the Gamma distributions. 
The distributions (\ref{lambda}), (\ref{gamma}), and (\ref{weight}) represented prior distributions, and (\ref{y}) represented the likelihood. 
SVGD approximated the target posterior distribution $p(\mathcal W, \lambda,\gamma|D)$ {to obtain the weights of the neural network.  In this sense, SVGD was used as an optimizer for the Bayesian neural network.}

\begin{figure}[t]
  \centering
  \includegraphics[width=0.95\columnwidth]{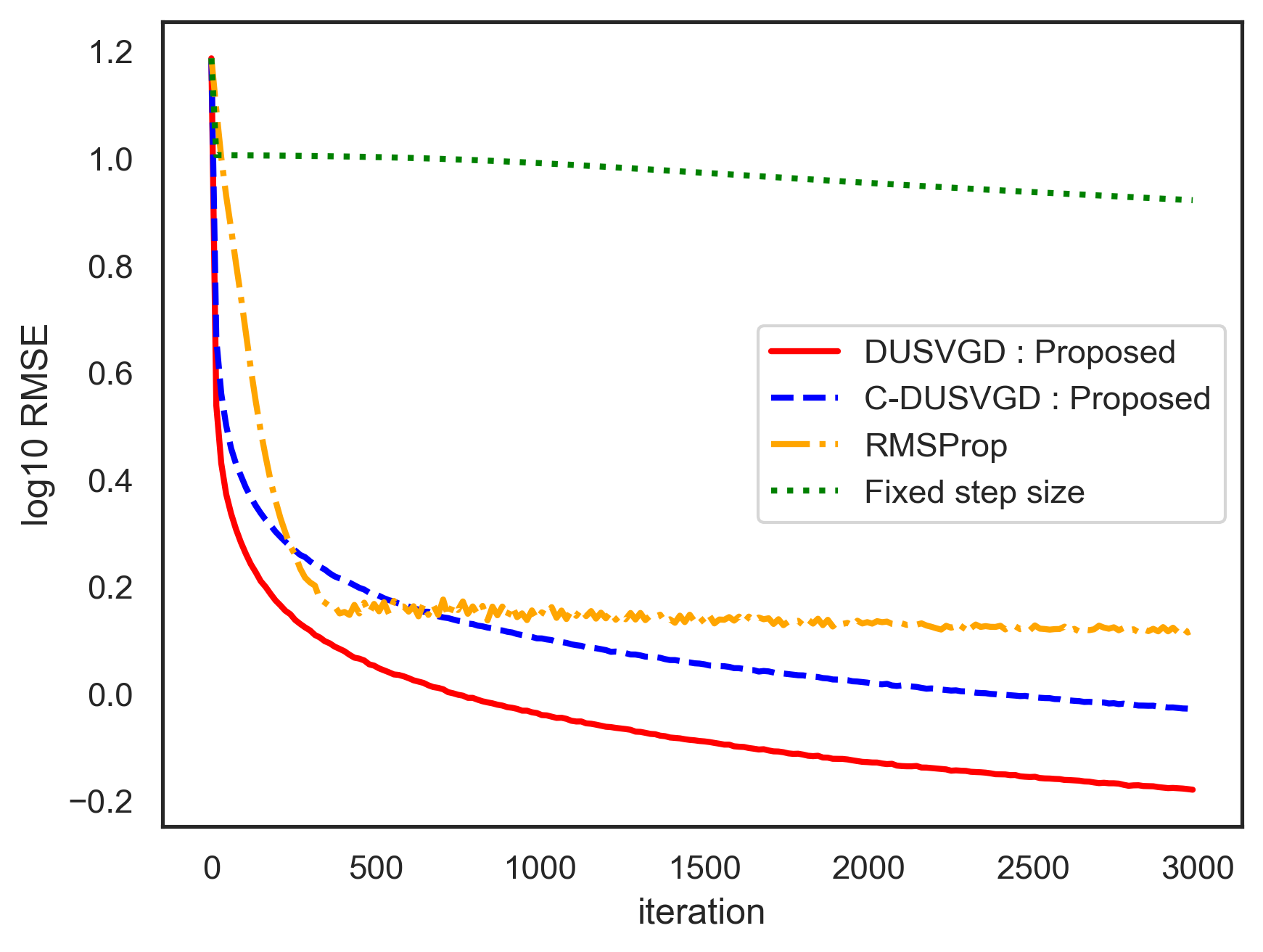}
  \caption{Dependency of the mean squared error on the number of iterations for DUSVGD, RMSProp, and fixed step size in Bayesian neural networks.}
  \label{fig:nn}
\end{figure}

{In the experiment, the values of hyperparameters were set to $a_{\lambda}=a_{\gamma}=1$ and $b_{\lambda}=b_{\gamma}=0.01$. 
In SVGD, $M=100$ particles were initialized by random numbers drawn from the prior distributions.}
The maximum number of iterations for learning (C-)DUSVGD was set to $T=15$, and the learned step sizes were used periodically.
{We used 90\% of $506$ data points for training and the remaining} {10\%} {for testing.}  The minibatch size was $B=1$, with $E=500$ learning epochs for DUSVGD and C-DUSVGD. {As a loss function, we used the mean squared error between the output $\bm{\hat y}$ and true $\bm{y}$.}
 The learning rate for the Adam optimizer was set to $1.0\times 10^{-6}$ for DUSVGD and $1.0$ for C-DUSVGD. 
 The initial value of trainable parameters were $\gamma_t = 1.0\times 10^{-4}$ for DUSVGD and $(\alpha,\beta)=(50,50)$ for C-DUSVGD.

Figure \ref{fig:nn} shows {the root mean squared error (RMSE) as a generalization loss, as a function of the number of iterations for DUSVGD, RMSProp, and fixed step size} in Bayesian neural networks. 
{RMSE was estimated by averaging over 10 trials. 
It was found that DUSVGD and C-DUSVGD improved the convergence speed and approximation performance compared to the other methods.} 
For example,  the value of $\log_{10}\mathrm{RMSE}$ after  $3000$ iterations was approximately $-1.8\times 10^{-1}$ for DUSVGD, $-2.8\times 10^{-2}$ for C-DUSVGD, $1.2\times 10^{-1}$ for RMSProp, and $9.2\times 10^{-1}$ for fixed step size.
In addition, we found that the RMSE of RMSProp stopped at approximately $1.2\times 10^{-1}$, implying that its particle distribution converged to a suboptimal distribution, possibly because of the  nonconvexity of the KL divergence minimized by SVGD. 
In contrast, the RMSE of (C-)DUSVGD continued decreasing after $3000$ iterations. This indicated that learning step size parameters could avoid convergence to undesired local minima.
Compared to that of C-DUSVGD, the RMSE performance of DUSVGD was reasonably better, suggesting that the performance depended on the flexibility of the unfolded algorithms.

{It is emphasized that training SVGD for optimizing Bayesian neural networks is regarded as meta-learning}{\cite{meta}}. {The proposed trainable SVGD algorithms show remarkable performance improvement, although they are trained with only $15$ iterations. 
The result implies that the proposed algorithms are promising meta-learning techniques for Bayesian neural networks.}

\section{Conclusion}

In this paper, we proposed DUSVGD and C-DUSVGD by applying DU to {SVGD, a representative ParVI algorithm for sampling a target distribution.
The numbers of trainable parameters of DUSVGD and C-DUSVGD were $T$ and $2$, respectively,} {both of which were independent of the dimension} {of a target distribution and the number of particles. 
These algorithms enabled us to tune step sizes flexibly via deep-learning techniques, according to the task and data set. 
In particular, C-DUSVGD  employed the theory of the Chebyshev step, resulting in the lowest number of trainable parameters. 
We conducted numerical simulations for three tasks to examine the proposed (C-)DUSVGD. 
For sampling from the Gaussian mixture distribution and Bayesian logistic regression problem, DUSVGD and C-DUSVGD improved the convergence speed compared to classical SVGD. Furthermore, the result of learning Bayesian neural networks indicated that training (C-)DUSVGD improved not only the convergence speed but also the learning performance. 
In the experiments, (C-)DUSVGD were learned with the first $10\mathrm{-}15$ iterations, and their learned parameters were used repeatedly. 
This greatly reduced the training costs of the proposed trainable SVGD, which was applicable to meta-learning.}
Future work may include applying DU to improve ParVI methods designed for accelerating convergence \cite{parvis}.

\end{document}